# The Golden Rule as a Heuristic to Measure the Fairness of Texts Using Machine Learning

Ahmed Izzidien, David Stillwell

## Abstract

In this paper we present a natural language programming framework to consider how the fairness of acts can be measured. For the purposes of the paper, a *fair* act is defined as one that one would be accepting of if it were done to oneself. The approach is based on an implementation of the golden rule (GR) in the digital domain. Despite the GR's prevalence as an axiom throughout history, no transfer of this moral philosophy into computational systems exists. In this paper we consider how to algorithmically operationalise this rule so that it may be used to measure sentences such as 'the boy harmed the girl' and categorise them as *fair* or *unfair*. A review and reply to criticisms of the GR is made. A suggestion of how the technology may be implemented to avoid *unfair* biases in word embeddings is made - given that individuals would typically not wish to be on the receiving end of an unfair act, such as racism, irrespective of whether the corpus being used deems such discrimination as praiseworthy.

## Introduction

The axiom of the Golder Rule (GR) to do onto others as one would wish upon oneself (Singer, 1963) has received little attention in the field of machine learning and information technology. Despite its simplicity, it potentially offers a unique method for evaluating whether an act, such as 'murder' or 'thanking', as described in a sentence, is one that is fair or unfair. In this paper, we focus on using this axiom to test straightforward test sentences describing a social interaction, such as 'the clerk murdered the prisoner', or 'the lawyer bribed the judge', i.e., a description of an act between an agent and a patient. We do so to explore both the technical feasibility of the approach, and philosophical justification for the method's use.

We build on earlier work by (Izzidien, 2022; Jentzsch et al., 2019; Schramowski et al., 2020), who use word embedding similarity-based approaches to classify terms as belonging to do/do-not words and moral-immoral classes.

The contrast and original contributions of this paper focus on a consideration of the validity of the philosophical approach of the GR, and the development of a technical method to use this philosophical assessment in the digital domain. A unique feature of using these methods, is that they are in no need of a list of do's and don'ts, which is a common approach in applying deontological and utilitarian ethics (Cervantes et al., 2020).

The paper also contributes originally through a philosophical discussion on the use of the GR for defining an act as fair or unfair. Lastly, we introduce an alternative perspective to thinking about fairness in AI.

The paper is split into two parts, a philosophical discussion of the GR, and a technical section.

## Section 1 Interrogating the Golden Rule

Our proposal of using the GR requires an investigation of its soundness as a moral philosophy. In this section, we consider the main criticisms made of the GR, and argue for its validity as a sound axiom.

A common criticism is given by Kant for the rule's seemingly being dependent upon one's personal taste. As George Bernard Shaw quipped "Don't do to others as you want them to do unto you. Their tastes may be different" (Shaw, 2008). This criticism may be replied to by observing that, in considering another person's wishes, one avoids this challenge. Thus, if John finds it amusing to be called a *yuppy*, he ought not use the same term to describe another person, if that other person does not find it amusing. In this manner, the rule can be understood as an invitation to duly consider any relevant difference between individuals - just as a person would wish such consideration from another (Wattles, 1997).

This approach can resolve many permutations of this objection. For example, a puzzle offered to readers in the opening chapter of Herman Melville's Moby-Dick presents Ishmael as being invited by his new friend, Queequeg, to join in pagan worship. The chapter has Ishmael consider: 'What do I wish that this Queequeg would do to me? Why, unite with me in my particular Presbyterian form of worship. Consequently, I must then unite with him in his; ergo, I must turn idolator' (Melville, 1950).

Had Ishmael abstracted to a higher level, he may have reached a different conclusion: 'What do I wish that this Queequeg would do to me? Why, unite with me in my particular Presbyterian form of worship. [Why so? Because I believe it to be a correct way of prayer. Consequently, I must invite him to a correct form of prayer, as I would wish others to invite me a to a correct form of prayer; ergo, I must invite him to Presbyterianism, and not turn to idolatry]'. Additionally, he may also consider the context negatively: [I would wish that others do not take me away from my correct form of prayer, ergo, I must remain a Presbyterianism, as I hold it to be correct]. By considering higher levels of abstraction, Ishmael avoids contradicting himself, since in all cases he would wish others to be Presbyterians, because he believes it to be true.

With another form of criticism, the wider implications of applying the GR are often left out. This may be illustrated by the following example. Were a criminal to be faced with prison, the criminal may suggest to the judge that they apply the GR. Whereby the criminal may claim that given that he would want to be let go, the judge should act according to what the criminal would want for himself in this context and set him free. However, in reply to this criticism, the judge may state that the criminal ought to apply the GR to himself, and consider that if he were a judge, he would not wish someone to ask him to break the law (Singer, 1963). Further, the judge would have to consider his act's wider implications, since in freeing the criminal, he is falling foul of the GR when applied to members of society who have a say in the decision. Others would not want criminals set free, therefor the judge ought not, according to the GR, impinge on their desire by setting criminals loose (Hare, 1977).

A further criticism of the GR is that applying the GR would lead an individual to be constantly at the mercies of the wishes of the recipients. They would always be doing what other people want. The person's own wishes and needs would go unmet. In reply, one may consider a society, in which everyone applied the GR, in which case the latter argument would not necessarily stand, since no-one would wish to be at the mercy of the wishes of others, and thus in applying the GR, ought to avoid imposing such an outcome on others. Furthermore, I would not want to have someone act towards me in a manner that was of detriment to their well-being, therefore I should abstain from acting in a manner that is detrimental to my wellbeing. Considering it at a higher level of abstraction also resolves this. Given that humans are averse to acts which do not bring some form of gain (Bussanich & Smith, 2013; Chislenko, 2020), the case above becomes: The receiver by virtue of their innate aversion to ungainful acts, would not want me to act in a manner that was ungainful to my well-being, since they would not wish that upon themselves. Thus, I ought not act in a manner that is not bringing me any gain but causing me great distress. In this manner I am considering their wishes before acting.

A further criticism levelled at the GR is the case of a sadomasochist. For example, two individuals agree to meet each week to violently pluck out a fingernail from each other. Both make the claim that they are adhering to the GR. A reply to this, is that one cannot claim humanity and dignity by appeal to the GR while at the same time committing inhumane and undignified acts, as such would be a contradiction in the premises behind both positions.

We expand on this as follows: two factors may be considered, the first, being their implicit acceptance of the concept of consent. Each is consenting to have their fingernails removed. This concomitantly implies their acceptance of the concept of non-consent. As well as personal autonomy from interference. Their acceptance of these implies, from their perspective, that humans may validly be attributable a meta-value that reflects human dignity and humanity. To them, these are valid ontological characterisations of humans. As a counter illustration, a stone, may be interfered with, indeed broken, without the actor being referred to as being inhumane and undignified to the stone. Humans, from the perspective of the two sadomasochists, possess qualities which imply humanity and dignity. Yet, humanity and dignity are not susceptible to suspension. A person may commit acts that are inhumane and undignified, yet, they remain human, deserving of humane and dignified treatment, even in punishment. If a person commits acts that impinge on their own dignity, though inhumane acts, they are committing the following contradiction:

1. A belief that humans have dignity and humanity.
2. Acting in a manner that contradicts that belief.

In violently torturing each other, we argue that their position is ineligible, since they are acting inhumanely, while holding the belief that humans ought to be treated humanely.

If they would wish to re-define humanity and dignity, they would have to consider the consequential effects of the act such a re-definition. Whereby a torturer may claim that the tortured person had consented to such punishment, even if denied by the victim. The consequence of normalising such inhumane acts has knock-on effects that can determine it to be in conflict with the very notion of human dignity that they may try to argue for based on a re-definition of an act. Furthermore, at a more abstract level, it may be argued that their implicit acceptance of the validity of redefining what humanity and dignity is, implies the validity of any other definition due to a lack of any ontological authority, as implied by their relativist position. Although it is beyond the remit of the paper to argue for natural moral realism, it may be considered that given that humans have evolved as social species (Nowak, 2006; Peysakhovich et al., 2014), they are inherently nudged into an *ought* that preserves society. Acts that involve going against the grain of social evolution at a fundamental level, such as defining violent torture as acceptable, are often redressed by the natural order of things, and are evolved out of the eco-system. Not that this biological propensity for pro-social acts defines a normative stance, but that the natural state of affairs points towards the upkeep of humane society (Tomasello, 2014). As such, the GR may be said to imply human dignity, and thus undignified acts would be inadmissible to avoid contradiction. While this argument relies on a reference point for human dignity, it rests on the definitions implicitly accepted by the agent, both direct and consequential. As a final point, it may also be argued, as with the case of the prior mentioned judge and wry criminal, the GR has a societal dimension, whereby the wishes of the wider society ought to be considered. In this manner, a consensus based on the lowest common denominator of what constitutes fair society becomes potentially realisable. This aspect of the GR may be said to operate within social contracts.

A final criticism, albeit hypothetical, was given by Rorty (Rorty, 1990), who presents what he considers to be the *consistent Nazi*. From the agent's perspective Rorty holds that there would be no way to refute a sophisticated, consistent, passionate Nazi psychopath who would favour his own elimination if he himself turned out to be Jewish. In reply, two factors may be considered, the earlier reply given to consensual violent torture, and second, that the perspective of the agent and premises for his identity as a Nazi have now changed. As such one cannot assume the views (i.e., Nazism) that were built on prior premises will remain.

While many categories of fairness exists such as procedural, distributional and interactional (Faullant et al., 2013), we employ the use of the GR as a means to capture fairness at a most basic level. That is, an act is fair, if I would be willing to accept it being done to myself; be that a procedural act, distributive act or interaction act.

**Using language models for the Golden Rule**

Word embeddings have been used to develop language models which can predict a missing word within a sentence. One such model is that of 'A Lite Bidirectional Encoder Representations from Transformers' (ALBERT) (Lan et al., 2020). Thus, a sentence such as: "London is the [MASK] of Britain", produces the result: 'capital' for the masked word.

This may be used to formulate the GR. For example, if a test sentence were to read, 'the man murdered the security officer', a GR reformulation - as given in the introduction, could read 'a man would [MASK] like to be murdered', for which ALBERT predicts: 'not'. Arguably the result is based on a human propensity to be averse to gainless harmful acts being reflected in the used corpus.

Once again this requires that the corpus that ALBERT is trained is representative of typical human society, i.e., not a corpus based on alternate realities, whereas previously mentioned, people enjoy being murdered.

A further consideration is in regards bias in datasets. It may be argued that where a corpus contains elements of praise for anti-social acts such as harming people due to their race, then the use of that corpus in a language model may replicate this bias. One way of avoiding this with the GR is to ask if the perpetrator would wish to be treated in this manner, to which the GR evaluation would be negative.

Thus, using this method potentially allows for a sentence can be classed as: *fair* or *unfair*, as given above in the 'Interrogating the Golden Rule' section.

ALBERT itself, is a transformer model pretrained on a large corpus of English data in a selfsupervised fashion, i.e., with no human labelling. Allowing it to use a large amount of publicly available data. It is pretrained with two objectives: Masked language modelling (MLM) and Sentence Ordering Prediction (SOP). With the former, the model randomly masks 15% of the words in the input then runs the complete masked sentence through the model in order to predict what the masked words are. This contrasts with traditional recurrent neural networks (RNNs) that usually see the words in sequence, or from autoregressive models like generative pre-trained transformers, which work by internally masking future tokens. This novelty allows for the model to learn a bidirectional representation of the sentence.

**Limitations**

## Punitively Fair Sentences and the GR

In general, people committing illegal acts do not wish to be arrested for their actions. This may pose an evaluative challenge to the current format of using language model masking. For example, "The police arrested the hacker", cannot be re-synthesised into a GR using the template: 'A hacker would [MASK] wish to be arrested'. As typically no one is happy being arrested. Using this template produces 'never'.

This limitation can be addressed by applying the theory on the GR given earlier. The sentence is formulated by considering: Would the person being arrested, or sanctioned in any way, for any act, wish the same act on themselves. Thus, the template becomes: 'A hacker is [MASK] happy being hacked", produces the result: 'never' (0.44).

## Further work:

### Rights and Duties

Given the ability to qualify and quantify social interactions in terms of their fairness – as defined in this paper, a definition which implicitly incorporates a determination of harm, and personal interest in being free from such harm, it becomes potentially possible to allocate individual and communal rights based on the well-established interest theory of rights and their corelative duties (Kramer, 2010).

Interest theory holds that a function of human rights is to protect and promote essential human interests (Tasioulas, 2015). According to Kramer's Bentham's test, a party holds a "right correlative to a duty only if that party stands to undergo a development that is typically detrimental if the duty is breached". Thus, a duty of that sort must be 'typically', 'normally' or 'standardly' in the interests of the duty bearer, otherwise the theory would be under-inclusive (Kramer, 2001, 2017). In being able to identify a harm to that interest using the GR, it becomes possible to allocate duties to avoid such harm. However, such an undertaking naturally requires separate and additional analysis, such as the detection of agreement clauses in a contract, or statute identification relevant to the context to begin the process of allocation. Use of the GR in this regard, with its explainable dimensions, may offer a new way of thinking about how such analytics and indeed ethics may be incorporated into a ML pipeline. Indeed, tort law is generally built on the ideas of harm, caused by wronging and rights violation. As such, downstream tasks and AI common sense reasoning may be able to better capture these dimensions for incorporation into their analysis.

A further avenue for the technology exists in using such analysis in visual artificial intelligence, whereby visually captured contexts may be translated into textual descriptions, from which a GR analysis may be performed.

Wallach (2010) has advocated for the building of 'moral machines' as a realistic goal, to ensure such autonomous machines do no harm humans. While others have advocated for the removal of such machines from such contexts (Sharkey, 2020), it would arguably be advantageous for an 'ethical machine' to be able to beat a 'malicious machine' to the finish line (Davies, 2016). Although robot ethics and machine morality publications have seen dramatic increase in number and readership, the relatively simple approach proposed in this paper was under reported.

Indeed one of the advantages of such an approach is that it does not set a list of rules to be fixed, posing questions of contextual inflexibility, and of questionable robustness for real-world tasks (Allen et al., 2005), but instead is able to promote a form of moral competence that people expect of one another (Malle, 2016).

This form of comparison has recently been considered by (Loi et al., 2020) as a means to overcome an over simplification in algorithmic fairness. Given the philosophy of the GR allows for individual circumstances and tastes to be considered, it's operationalization in these domains has the potential of covering a gap which exists between the two different concepts of discrimination: comparative and non-comparative (Hellman, 2020). The comparative notion of discrimination determines if an outcome is fair by making reference to other people. Yet, with a non-comparative view of discrimination, being fair consists in treating each individual as they are entitled to be treated (Hellman, 2016). This incorporates a consideration of the particular features and characteristics of the individual, and is a far cry from many current algorithmic fairness methods that purposefully mask such characteristics (Loi et al., 2020).

## Concluding Remarks

In this paper we proposed a novel incorporation of a longstanding moral philosophy into machine learning. The general flexibility of the rule, the Golden Rule, makes its arguably ideal for cross-cultural assessments. A number of implementations using word embeddings were proposed, to include its use in legal rights and duties allocation, non-comparative algorithmic fairness and as a heuristic for the development of ethical general intelligence.

## Acknowledgments

AI designed the study, coded the software, analysed the results and wrote the manuscript. DS commented on the paper and overall approach used.